\documentclass[]{article}

\usepackage{hyperref}
\hypersetup{
	pdftitle={Translation-Invariant Shrinkage/Thresholding of Group Sparse Signals}, 
	pdfauthor={Po-Yu Chen and Ivan Selesnick},
	colorlinks=true,
	linkcolor = black,
	citecolor = black,
	urlcolor = blue,
	breaklinks = true,
}

\usepackage{amssymb}
\usepackage{amsmath}
\usepackage{booktabs}
\usepackage{graphicx}
\usepackage{epstopdf}
\usepackage{cite}
\usepackage{verbatim}
\usepackage{url}

\usepackage{mathtools}
\usepackage{array}
\usepackage{centernot}

\DeclareMathOperator{\soft}{soft}

\DeclareMathOperator{\erf}{erf}

\providecommand{\norm}[1]{\lVert#1\rVert}
\providecommand{\abs}[1]{\lvert#1\rvert}

\newcommand{\myE}{ \mathrm{e} }

\newcommand{\lam}{{\lambda} }
\newcommand{\half}{\frac{1}{2}}

\newcommand{\ct}{^{\ast}}

\newcommand{\s}{\mathbf{s}}

\newcommand{\ub}{\mathbf{u}}
\renewcommand{\v}{\mathbf{v}}
\newcommand{\w}{\mathbf{w}}
\newcommand{\x}{\mathbf{x}}
\newcommand{\y}{\mathbf{y}}

\newcommand{\I}{\mathbf{I}}

\newcommand{\0}{\mathbf{0}}

\newcommand{\calN}{\mathcal{N}}

\newcommand{\calJ}{\mathcal{J}}
\newcommand{\calI}{\mathcal{I}}

\newcommand{\CC}{\mathbb{C}}
\newcommand{\RR}{\mathbb{R}}

\newcommand{\ia}{({\it i\/})}
\newcommand{\ib}{({\it ii\/})}
\newcommand{\ic}{({\it iii\/})}

\newcommand{\cf}{F}				
\newcommand{\tr}{^{t}}			
\newcommand{\opt}{^{\ast}}
\newcommand{\NZ}{\calI'}		
\DeclareMathOperator{\ogs}{ogs}

\usepackage[vmargin = 0.9in, hmargin = 0.97in]{geometry}

\usepackage{setspace}
\newcommand{\prog}[1]
{
\noindent
  \begin{footnotesize}
  \renewcommand{\baselinestretch}{1.2}
  \verbatiminput{#1}
  \end{footnotesize}
}

\title{Translation-Invariant Shrinkage/Thresholding of Group Sparse Signals%
\thanks{Corresponding author: Ivan W. Selesnick.}%
\thanks{This work is supported by the NSF under Grant No. CCF-1018020.}%
}
\author{Po-Yu Chen and Ivan W. Selesnick\\
Polytechnic Institute of New York University\\ 6 Metrotech Center, Brooklyn, NY 11201, USA\\
Email: poyupaulchen@gmail.com, selesi@poly.edu. Tel: +1 718 260-3416.%
}
\date{}

\renewcommand\footnotemark{}

\begin{document}
\maketitle

\begin{abstract}
This paper addresses signal denoising when large-amplitude coefficients form clusters (groups).
The L1-norm and other separable sparsity models do not capture the tendency of coefficients to cluster (group sparsity).
This work develops an algorithm, called `overlapping group shrinkage' (OGS), based on the minimization
of a convex cost function involving a group-sparsity promoting penalty function.
The groups are fully overlapping so the denoising method is translation-invariant and blocking artifacts are avoided.
Based on the principle of majorization-minimization (MM),
we derive a simple iterative minimization algorithm that reduces the cost function monotonically.
A procedure for setting the regularization parameter,
based on attenuating the noise to a specified level, is also described. 
The proposed approach is illustrated on speech enhancement, 
wherein the OGS approach is applied in the 
short-time Fourier transform (STFT) domain.
The denoised speech produced by OGS does not suffer from musical noise.
\end{abstract}

\section{Introduction}

In recent years, many algorithms based on sparsity have been developed for signal denoising,
deconvolution, restoration, and reconstruction, etc.\ \cite{Elad_2010_pieee}.
These algorithms often utilize nonlinear scalar shrinkage/thresholding functions of various forms
which have been devised so as to obtain sparse representations.
Examples of such functions are the
hard and soft thresholding functions \cite{Don95},
and
the nonnegative garrote \cite{Gao_1998, Figueiredo_2001_TIP}.
Numerous other scalar shrinkage/thresholding functions have been derived 
as MAP or MMSE estimators using various probability models, e.g.\  \cite{Hyvarinen_1999,Fadili_2005_TIP,Mallat_1989}.

We address the problem of denoising, i.e. estimating $x(i)$, $ i \in \calI $, from noisy observations $ y(i) $, 
\begin{equation}
	\label{eq:yaw}
	y(i) = x(i) + w(i), 
	\qquad
	i \in \calI,
\end{equation}
where the signal $ x(i) $ is known to have a group sparse property and $w(i)$ is white Gaussian noise.
Here, $ \calI $ denotes the domain of $ x $, typically $ \calI = \{0, \ \dots, \ N-1 \} $ for one-dimensional finite-length signals.
A generally effective approach for deriving shrinkage/thresholding functions
is to formulate the optimization problem
\begin{equation}
	\label{eq:opt}
	\x\opt = \arg \min_{\x} \; \left\{ F(\x) = \frac{1}{2} \, \norm{\y - \x}_2^2 +  \lam R(\x)   \right\}
\end{equation}
where $ \x = (x_i, \, i \in \calI) $ is the signal to be determined
from the observation $ \y = (y_i, \, i \in \calI)$.
In problem \eqref{eq:opt}, 
$\x$ may represent either the coefficients
of a signal (e.g.\ wavelet or short-time Fourier transform coefficients) or the signal itself, if the signal is well modeled as sparse. 
The penalty function $R(\x)$ (regularizer) should be chosen 
to promote the known behavior of $ \x $.
Many of the shrinkage/thresholding functions devised in the literature
can be derived as solutions to \eqref{eq:opt},
where $R(\x)$ is specifically of the separable form
\begin{equation}
	\label{eq:reg}
	R(\x) = \sum_{i \in \calI} r(x(i)).
\end{equation}
For example, soft-thresholding is derived as the solution to \eqref{eq:opt} where $r(x) = \abs{x} $,
which corresponds to 
the lasso problem \cite{Tibshirani_1994} or the basis pursuit denoising problem \cite{Chen_SIAM_1998}.
When $R(\x)$ has the form \eqref{eq:reg}, the variables $x(i)$ in \eqref{eq:opt}
are decoupled, and the optimization problem \eqref{eq:opt} is equivalent to a set
of scalar optimization problems.
Therefore, the separable form \eqref{eq:reg} significantly simplifies the task of solving \eqref{eq:opt},
because in this case the optimal solution is obtained by applying a scalar shrinkage/thresholding function
independently to each element $x(i)$ of $\x$.
From a Bayesian viewpoint, the form \eqref{eq:reg} models the elements $x(i)$
as being statistically independent. 

For many natural (physically arising) signals, the variables (signal/coefficients) $\x$ are not only sparse
but also exhibit a clustering or grouping property.
For example, wavelet coefficients generally have inter and intra-scale clustering tendencies
\cite{Liu_2001_TIP,Shapiro_1993_TSP,Simoncelli_2001}.
Likewise, the clustering/grouping property is also apparent in a typical speech spectrogram.
In both cases, significant (large-amplitude) values of $\x$ tend not to be isolated.

This paper develops a simple translation-invariant shrinkage/thresholding algorithm
that exploits the grouping/clustering properties of the signal/coefficient vector $\x$.
The algorithm acts on $\x$ as a whole without performing block-by-block processing,
and minimizes the cost function \eqref{eq:opt}
with the (non-separable)  penalty function
\begin{equation}
	\label{eq:ogcost}
	R(\x) = \sum_{i \in \calI} \left[ \sum_{j \in \calJ} \abs{x(i+j)}^2 \right]^{1/2},
\end{equation}
where the set $ \calJ $ defines the group.
The algorithm is derived using the majorization-minimization (MM) method \cite{FBDN_2007_TIP, Oliveira_2009_SP}.
The algorithm
can be considered an extension of the successive substitution algorithm for multivariate thresholding \cite{SenSel-02-tsp}
or as an extension of the FOCUSS algorithm \cite{Rao_2003_TSP}.

The penalty function \eqref{eq:ogcost} has been considered previously,
but existing algorithms for its minimization
\cite{Jenatton_2011_MLR, Deng_2011_techrep, Figueiredo_2011_SPARS, Bayram_2011_picassp, Peyre_2011_eusipco, Bach_2012_techrep, Jenatton_2010_ICML} are based on variable duplication
(variable splitting, latent/auxilliary variables, etc.).
The number of additional parameters is proportional to the overlap;
hence their implementations require additional memory  and accompanying data indexing.
The iterative algorithm proposed here avoids variable duplication and can be efficiently implemented
via separable convolutions.

For the purpose of denoising,
this paper also develops a conceptually simple method to set the regularization parameter $\lambda$
analogous to the `3$\sigma$' rule.
The method allows for $\lambda$ to be selected so as to ensure that the noise variance
is reduced to a specified fraction of its original value.
This method does not aim to minimize the mean square error or any other measure
involving the signal to be estimated,
and is thus non-Bayesian.
Although conceptually simple, the method for setting $\lambda$ is analytically intractable 
due to the lack of explicit functional form of the estimator.
However, with appropriate pre-computation, the method can be implemented by table look-up.

In Section \ref{sec:algo}, the cost function is given and the iterative successive substitution algorithm 
for its minimization is presented.
In Section \ref{sec:pnoise}, the effect of the shrinkage/thresholding algorithm on white Gaussian noise
is investigated and is used to devise a simple method for selecting the regularization parameter $\lam$.
Section \ref{sec:examples} illustrates the proposed approach to signal denoising, including speech enhancement.

\subsection{Related work}

The penalty function \eqref{eq:ogcost} can be considered a type of mixed norm.
Mixed norms and their variants can be used to describe non-overlapping group sparsity
\cite{Yuan_2006, Kowalski_2009_SIVP, Kowalski_2009_ACHA, Eldar_2009_Tinfo}
or overlapping group sparsity
\cite{Jenatton_2011_MLR, Deng_2011_techrep, Figueiredo_2011_SPARS, Jacob_2009_cnf, Bayram_2011_picassp, Peyre_2011_eusipco, Bach_2012_techrep, Mosci_2010_NIPS, Yuan_2011_NIPS, Chen_2012_AAS}.
Algorithms derived using 
variable splitting and the alternating direction method of multipliers (ADMM) 
are described in \cite{Deng_2011_techrep, Figueiredo_2011_SPARS, Boyd_2011_admm}.
These algorithms duplicate each variable for each group of which the variable is a member.
The algorithms have guaranteed convergence, 
although additional memory is required due to the variable duplication (variable splitting).
A theoretical study regarding recovery of group support is given in \cite{Jacob_2009_cnf} which uses an algorithm that is also based on 
variable duplication.
A more computationally efficient version of \cite{Jacob_2009_cnf} for large data sets
is described in \cite{Mosci_2010_NIPS} which is based on identifying active groups (non-zero variables). 
The identification of non-zero groups is also performed in \cite{Yuan_2011_NIPS} using an
iterative process to reduce the problem size;
a dual formulation is then derived which also involves auxiliary variables.
Auxiliary and latent variables are also used in the algorithms described in \cite{Peyre_2011_eusipco, Bach_2012_techrep, Bayram_2011_picassp,Jenatton_2010_ICML},
which, like variable splitting, calls for additional memory proportional to the extent of the overlap.
Overlapping groups are used to induce sparsity patterns in the wavelet domain in \cite{Rao_2011_ICIP}
also employing variable duplication. 

Several other algorithmic approaches and models have been devised to
take into account clustering or grouping behavior,
such as: Markov models \cite{CNB98, Pizurica_2002_TIP, Gehler_2005,Molina_2010_GRLS, Dikmen_2010_TASLP, Fevotte_2008_TSALP, Wolfe_2004_royal,Chaari_2010_TSP},
Gaussian scale mixtures  \cite{Portilla_2003_TIP, Goossens_2009_TIP},
neighborhood-based methods with locally adaptive shrinkage \cite{Cai_2001, Shui_2007, MIRM99},
and
multivariate shrinkage/thresholding functions \cite{SenSel-02-tsp, Sendur_2002_asilomar, Rabbani_2009_TBME, Achim_2005_SPL, Cho_2005,Pustelnik_2011_TIP,Selesnick_2008_TSP_MLap}.
These algorithms depart somewhat from the variational approach wherein a cost function 
of the form \eqref{eq:opt} is minimized.
For example, in many neighborhood-based and multivariate thresholding methods,
local statistics are computed for a neighborhood/block around each coefficient, and then these statistics
are used to estimate the coefficients in the neighborhood (or the center coefficient).
In some methods, the coefficients are segmented into non-overlapping blocks and each block
is estimated as a whole; however, in this case the processing is not translation-invariant
and some coefficient clusters may straddle multiple blocks.

It should be noted that an alternative penalty function for promoting group sparsity is proposed in \cite{Obozinski_2011_techrep, Jacob_2009_cnf}
(see equation (3) of \cite{Obozinski_2011_techrep}).
Interestingly, it is designed to promote sparse solutions for which the significant values tend to be comprised of
the unions of groups;
while, as discussed in \cite{Obozinski_2011_techrep, Jacob_2009_cnf}, the penalty function \eqref{eq:ogcost}
promotes sparse solutions for which the significant values tend to be comprised of
the \emph{complement} of the unions of groups.
The focus of this manuscript is an efficient algorithm for
the minimization of \eqref{eq:opt} with penalty function \eqref{eq:ogcost} and
the corresponding selection of regularization
parameter $\lambda$.
However, the extension of this work to the penalty function of \cite{Obozinski_2011_techrep, Jacob_2009_cnf}
is also of interest.

\section{Overlapping group shrinkage/thresholding}
\label{sec:algo}

\subsection{Motivation}

As shown in \cite{CNB98, Liu_2001_TIP,Simoncelli_2001},
neighboring coefficients in a wavelet transform have statistical dependencies even when
they are uncorrelated.
In particular, a wavelet coefficient is more likely to be large in amplitude if the adjacent coefficients
(in scale or spatially) are large.
This behavior can be modeled using suitable non-Gaussian multivariate probability density functions,
perhaps the simplest one being
\begin{equation}
	\label{eq:bigroup}
	p(\x) = \frac{C}{\sigma^d} \, \exp{\left( -\frac{\sqrt{d+1}}{\sigma} \norm{\x}_2 \right)},
	\quad \x \in \RR^d
\end{equation}
as utilized in \cite{SenSel-02-tsp}.
If the coefficients $\x$ are observed in additive white Gaussian noise,
$\y = \x + \w$, 
then the MAP estimator of $\x$ is obtained by solving \eqref{eq:opt}
with $ R(\x) = \norm{\x}_2 $,
the solution of which is given by
\begin{equation}
	\label{eq:bisoft}
	\hat{\x} = \left( 1 - \frac{\lambda}{\norm{\y}}  \right)_{+} \, \y,
\end{equation}
where $(x)_{+} := \max(x, 0) $.
The function \eqref{eq:bisoft} can be considered a multivariate form of soft thresholding
with threshold $\lambda$.

The multivariate model and related models are convenient
for estimating small blocks/neighborhoods within a large array of coefficients; however, 
when each coefficient is a member of multiple groups, then 
either estimated coefficients are discarded (e.g.\ only the center of each block
of estimated coefficients is retained as in sliding window processing)
or the blocks are sub-sampled so that they are not overlapping.
In the first case, the result does not correspond directly to the minimization 
of a cost function; 
whereas, in the second case, the process is not translation-invariant
and issues may arise due to blocks not aligning with the
group sparsity structure within the array.
Here we are interested in a variational approach based on
fully-overlapping groups so that the derived algorithm is translation-invariant
and blocking artifacts are avoided.
The penalty function \eqref{eq:ogcost} is clearly translation-invariant.

\subsection{Penalty function}

Assuming that $\x$ has a group sparsity (clustering) property, 
to perform translation-invariant denoising, 
we add the $\ell_2$ norm for each group
to obtain the penalty function \eqref{eq:ogcost}.
The penalty function \eqref{eq:ogcost} is convex
and the cost function $ F $ in \eqref{eq:opt} is strictly convex
(because $\norm{\y - \x}_2^2$ is strictly convex in $\x$).
The index $i$ is the group index, and $j$ is the coefficient index within group $i$.
Each group has the same size, namely $\abs{\calJ}$.
In practice, $\x$ is of finite size; hence, the sum over $i$ in \eqref{eq:ogcost} is a finite sum.
To deal with the boundaries of $ \x $, we take $x(i)$ in \eqref{eq:ogcost} as zero when $i$ falls outside $\calI$;
that is, $x(i) = 0$ for $ i \notin \calI $.

For one-dimensional vectors $\x$ of length $N$ with group size $K$, we set $\calI$ in \eqref{eq:yaw} to
\begin{equation}
	\label{eq:cali}
	\calI = \{0, \, \dots, \, N-1\},
\end{equation}
and $\calJ$ in \eqref{eq:ogcost} to
\begin{equation}
	\label{eq:calj}
	\calJ = \{0, \, \dots, \, K-1\}.
\end{equation}
Note that in \eqref{eq:ogcost}, the groups are fully overlapping,
as per a sliding window shifted a single sample at a time.
It is possible to include a weighting $w(j)$ in \eqref{eq:ogcost}
as in Ref.~\cite{Siedenburg_2011_DAFx, Siedenburg_2013_JAES}.

For a two-dimensional array $\x$ of size $ N_1 \times N_2 $ with group size $ K_1 \times K_2$,
we set $\calI$ in \eqref{eq:yaw} to
\[
	\calI = \{(i_1, i_2) \, : \, 0 \le i_1 \le N_1-1, \; 0 \le i_2 \le N_2-1 \},
\]
and $\calJ$ in \eqref{eq:ogcost} to
\[
	\calJ = \{(j_1, j_2) \, : \, 0 \le j_1 \le K_1-1, \; 0 \le j_2 \le K_2-1 \}.
\]
In the two-dimensional case, $ i + j $ denotes $ (i_1+j_1, i_2+j_2) $.
The same notation extends to higher dimensions.

Note that minimizing \eqref{eq:opt} with  \eqref{eq:ogcost} can only shrink the data vector $\y$ 
toward zero. 
That is, the minimizer $\x$ of \eqref{eq:opt} will 
lie point-wise between zero and $\y$, 
i.e.\  $ x(i) \in [0, y(i)] $ for all $ i \in \calI $.
This can be shown by observing that the penalty function in \eqref{eq:ogcost}
is a strictly increasing function of $\abs{x(i)}$ and 
is independent of the sign of each $x(i)$.
As a result, if $y(i) = 0$ for some $ i \in \calI$, then $x\opt(i) = 0$ also,
where $ \x\opt $ is the minimizer of $ F $.

An important point regarding $R$ in \eqref{eq:ogcost} is that it is non-differentiable.
In particular, if any group of $\x$ is equal to zero, then $R$ is non-differentiable at $\x$.

\subsection{Algorithm}

To minimize the cost function $\cf(\x)$ in \eqref{eq:opt}, we use the majorization-minimization (MM) method \cite{FBDN_2007_TIP}.
To this end, we use the notation
\begin{equation}
	\v_{i,K} = [v(i), \, \dots, v(i+K-1)]\tr \in \CC^K
\end{equation}
to denote the $i$-th group of vector $\v$.
Then the penalty function in \eqref{eq:ogcost} can be written as
\begin{equation}
	\label{eq:raik}
	R(\x) = \sum_{i} \norm{\x_{i,K}}_2.
\end{equation}
To majorize $R(\x)$, first note that
\begin{equation}
	\label{eq:bineq}
	\frac{1}{2} \left[ \frac{\norm{\v}_2^2}{\norm{\ub}_2}  + \norm{\ub}_2 \right] \ge \norm{\v}_2
\end{equation}
for all $ \v $ and $ \ub \neq \0 $,
with equality when $ \v = \ub $.
The inequality \eqref{eq:bineq} can be derived from 
$ t^2 + s^2 \ge 2 t s, \forall t, s \in \RR $
by setting $ t = \norm{\v}_2 $, $ s = \norm{\ub}_2$.
Define
\begin{equation}
	\label{eq:gdef}
	g(\x, \ub) = \half \sum_i \left[ \frac{\norm{\x_{i,K}}_2^2}{\norm{\ub_{i,K}}_2}  + \norm{\ub_{i,K}}_2 \right].
\end{equation}
If $\norm{\ub_{i,K}}_2 > 0$ for all $ i \in \calI $,
then it follows from  \eqref{eq:raik} and \eqref{eq:bineq}
that
\begin{align}
	g(\x, \ub) & \ge R(\x)
	\\
	g(\ub, \ub) & = R(\ub),
\end{align}
for all $ \x, \ub $. 
Therefore, $ g(\x, \ub) $ is a majorizer of $ R(\x) $ 
provided that $ \ub $ has no groups equal to zero.
Moreover, the elements of $ \x $ are decoupled in $ g(\x, \ub) $
and so  $ g(\x, \ub) $ can be written as
\begin{equation}
	 g(\x, \ub) = \half \sum_{i \in \calI} r(i; \, \ub) \abs{x(i)}^2 + c
\end{equation}
where
\begin{equation}
	r(i; \, \ub) := 
	\sum_{j \in \calJ}
	\left[
		\sum_{k \in \cal J}
		\abs{u(i-j+k)}^2
	\right]^{-1/2}
\end{equation}
and $ c $ is a constant that does not depend on $\x$.
A majorizer of the cost function $ \cf(\x) $ is now given by
\begin{equation}
	G(\x, \ub) = \frac{1}{2} \, \norm{\y - \x}_2^2 +  \lam \, g(\x, \ub).
\end{equation}
The MM method produces the sequence $\x^{(k)}$, $ k \ge 1 $, given by:
\begin{align}
	\x^{(k+1)} & = \arg \min_{\x} \; G(\x, \x^{(k)})
	\\
	\label{eq:updateA}
	& = \arg \min_{\x} \;  \norm{\y - \x}_2^2 + \lam \sum_{i \in \calI} r(i; \, \x^{(k)}) \, \abs{x(i)}^2
\end{align}
where $\x^{(0)}$ is the initialization.
Note that \eqref{eq:updateA} is separable in $ x(i) $, 
so we can write \eqref{eq:updateA} equivalently as
\begin{equation}
	\label{eq:updateAs}
	x^{(k+1)}(i) = \arg \min_{x \in \CC} \;  (y(i) - x)^2 +  \lam \, r(i; \, \x^{(k)}) \, \abs{x}^2.
\end{equation}
However, the term $r(i; \x^{(k)})$ is undefined if $ \x^{(k)}_{i,K} = \0 $,
i.e.\ if the $i$-th group is all zero.
This is a manifestation of the singularity issue which may arise whenever
a quadratic majorizer is used for a non-differentiable function \cite{FBDN_2007_TIP}.
Hence, care must be taken to define an algorithm that avoids operations involving undefined quantities.

Consider the following algorithm.
Define $ \NZ $ as the subset of $ \calI $ where $ x^{(0)}(i) \neq 0 $,
\begin{equation}
	\label{eq:ynz}
	\NZ := \{ i \in \calI \; : \; x^{(0)}(i) \neq 0 \}.
\end{equation}
Define the update equation by:
\begin{equation}
	\label{eq:updatea}
	x^{(k+1)}(i) =
	\begin{cases}
		\displaystyle
		\frac{y(i)}{1 + \lam \, r(i; \, \x^{(k)})},		\qquad	&	i \in \NZ
	\\
		0	& i \notin \NZ
	\end{cases}
\end{equation}
with initialization $\x^{(0)} = \y$.
Note that the first case of \eqref{eq:updatea} is the solution to \eqref{eq:updateAs}.
The iteration \eqref{eq:updatea} is the  `overlapping group shrinkage' (OGS) algorithm,
summarized in Table \ref{table:ogs} in the form of pseudo-code.

Several observations can be made.

It is clear from \eqref{eq:ynz} and \eqref{eq:updatea}, that
\begin{equation}
	x^{(0)}(i) = 0
	\quad \implies \quad	
	x^{(k)}(i) = 0, \; \text{for all $k \ge 1$}.
\end{equation}
Therefore, any values initialized to zero will equal zero throughout the course of the algorithm.
However, 
if $ x^{(0)}(i) = 0 $, then $ y(i) = 0 $ as per the initialization.
Note that if $ y(i) = 0 $, then the optimal solution $ \x\opt $ has $ x\opt(i) = 0 $.
Therefore
\begin{equation}
	x^{(0)}(i) = 0
	\quad \implies \quad	
	x^{(k)}(i) = x\opt(i), \; \text{for all $k \ge 1$}.
\end{equation}
As a side note, if $ \y $ is a noisy data vector, then it is unlikely that $ y(i) = 0 $ for any $ i \in \calI $.

Now consider the case where $ x^{(0)}(i) \neq 0 $.
Note that if $ x^{(k)}(i) \neq 0 $, then 
no groups of $\x^{(k)}$ containing $ x^{(k)}(i) $ are all zero.
Hence
$  r(i; \, \x^{(k)})  $ is well defined with $  r(i; \, \x^{(k)}) > 0 $.
Therefore $y(i) / [1 + \lam \, r(i; \, \x^{(k)})] $
is well defined, lies strictly between zero and $y(i)$, and has the same sign as $y(i)$.
Hence, by induction,
\begin{equation}
	\label{eq:ngtz}
	x^{(0)}(i) \neq 0
	\quad \implies \quad	
	x^{(k)}(i) \neq 0, \; \text{for all $k \ge 1$}.
\end{equation}
Therefore, any value not initialized to zero will never equal zero in any subsequent iteration.
However,  for some $ i \in \NZ $,  $x^{(k)}(i)$ may approach zero in the limit as $ k \rightarrow \infty $.
That is,
\begin{equation}
	x^{(0)}(i) \neq 0
	\quad \centernot\implies \quad	
	\lim_{k \to \infty} x^{(k)}(i) \neq 0.
\end{equation}
In the example in Sect.~\ref{sec:onedimeg}, 
it will be seen that some values do go to zero as the algorithm progresses.
In practice, \eqref{eq:ngtz} may fail for some $ i \in \NZ $ due to finite numerical precision.
In this case, such $ i $ should be removed from $ \NZ $ to avoid potential
`division-by-zero' in subsequent iterations.

\begin{table}
	\caption{
		\label{table:ogs}
		Overlapping group shrinkage (OGS) algorithm for minimizing \eqref{eq:opt} with penalty function \eqref{eq:ogcost}.
	}
\begin{center}
\begin{minipage}{2.7in}
\hrule
\begin{align*}
	&	\text{Algorithm OGS}
	\\
	&	\text{input: $\y \in \RR^N $, $ \lambda > 0 $, $ \cal J $}
	\\
	&
		\x = \y	\qquad \quad \text{(initialization)}
	\\
	&
		\NZ =  \{ i \in \calI, \ x(i) \neq 0 \}  
	\\
	& \text{repeat}
	\\
	&	\qquad
			r(i; \, \x) = \sum_{j \in \calJ}
			\left[	\sum_{k \in \cal J} \abs{x(i-j+k)}^2 \right]^{-1/2} \!,
			\ i \in \NZ
	\\[1em]
	&	\qquad
			x(i) = \frac{y(i)}{1 + \lam \, r(i; \, \x)}, \quad i \in \NZ 
	\\
	& \text{until convergence}
	\\
	& \text{return: $\x$}
\end{align*}
\hrule
\end{minipage}
\end{center}
\end{table}

The OGS algorithm produces sparse solutions
by gradually reducing non-zero values of $\y$ toward zero,
rather than by thresholding them directly to zero on any iteration,
as illustrated in Fig.~\ref{fig:example1_conv} below.

The output of the OGS algorithm will be denoted as $ \x = \ogs(\y; \, \lambda, K) $,
where $K$ is the block size.
The OGS algorithm can also be applied to multidimensional data $\y$ 
using the above multi-index notation, with
group size
$ K = (K_1, \dots, K_d) $
and where
 $ \calI $ and $ \calJ $ are multi-indexed sets as described above.

Note that when the group size is $ K  = 1 $, then 
$ \calJ $ in \eqref{eq:calj} is given by $ \calJ = \{0\} $, and
the penalty function \eqref{eq:ogcost}
is simply the $\ell_1$ norm of $\x$.
In this case, the solution is obtained by soft thresholding.
When the group size $K$ is greater than one, the groups overlap and every element of the solution $\x$
depends on every element of $\y$;
hence, it is not possible to display a multivariate
shrinkage function as in 
\cite{SenSel-02-tsp} for the non-overlapping case.

\smallskip
\noindent
\textbf{Implementation.}
The quantity $ r(i; \, \x)$ in step (2) of the OGS algorithm can be computed
efficiently using two convolutions --- one for the inner
sum and one for the outer sum.
The inner sum 
can be computed as the convolution of $\abs{x(\cdot)}^2$
with a `boxcar'  of size $\abs{\calJ}$.
Denoting by $g(\cdot)$ the result of the inner sum,
the outer sum is again a running sum
or `boxcar' convolution applied to $g(\cdot)^{-1/2}$.
In the multidimensional case, each of the two convolutions are
multidimensional but separable and hence computationally efficient.

The computational complexity of each iteration of the OGS algorithm is of order $ K N $,
where $ K$ is the group size and $N$ is the total size of $\y$.
The memory required for the algorithm is $ 2 N + K $.

\smallskip
\noindent
\textbf{Convergence.}
For the OGS algorithm, due to its derivation using majorization-minimization (MM), it is guaranteed that the cost function $F(\x)$ monotonically decreases from one iteration to the next, i.e.,
$ F(\x^{(k+1)}) < F(\x^{(k)})$ if $\x^{(k)} \neq \x\opt$.
Yet, the proof of its convergence to the unique minimizer is complicated by the `singularity issue' which arises when a quadratic function is used as a majorizer of a non-differentiable function \cite{FBDN_2007_TIP}.
For the OGS problem it is particularly relevant since, as in most sparsity-based denoising problems, $F(\x)$ will usually be non-differentiable at the minimizer $\x\opt$.
Specifically, for OGS, if $\x\opt$ contains any group that is all zero, then $F(\x)$ is non-differentiable at $\x\opt$.
However, several results regarding the singularity issue are given in \cite{FBDN_2007_TIP} which strongly suggest that this issue need not hinder the reliable convergence of such algorithms in practice.

For the OGS algorithm with the initialization $ \x^{(0)} = \y $, the component $\x^{(k)}(i)$ will never equal zero except when $y(i)$
itself equals zero, as noted above.
In the OGS algorithm, the singularity issue is manifested 
by $ r(i, \x^{(k)}) $ approaching infinity for some $ i \in \NZ $.
In particular, if $x\opt(i) = 0$ for some $i \in \NZ$, then $ r(i, \x\opt) $ is undefined (due to `division-by-zero').
For $i \in \NZ$ such that $ x\opt(i) = 0 $, the value of
$ r(i, \x^{(k)}) $ goes to infinity as the algorithm progresses,
and $ x^{(k)}(i) $ goes to zero.
Note that in the OGS algorithm, the term $ r(i, \x^{(k)}) $ is used
only in the expression $y(i) / [1 + \lam \, r(i; \, \x^{(k)})] $.
Therefore, even when $ r(i; \, \x^{(k)}) $ is very large, this expression is 
still numerically well behaved.
(When large values of opposite signs are added to obtain small numbers,
the result is not numerically reliable, but that is not the case here.)
If the algorithm is implemented in fixed point arithmetic, then it is indeed
likely that the large values of $ r(i; \, \x^{(k)}) $ will lead to overflow
for some $i \in \NZ$.
In this case, $ x^{(k)}(i) $ should be set to zero and $ i $ 
should be removed from $ \NZ $ for the subsequent iterations.

We do not prove the convergence of the OGS algorithm due to the singularity issue.
However, in practice, the singularity issue does not appear to impede the convergence of the algorithm,
consistent with the results of \cite{FBDN_2007_TIP}.
We have found that the empirical asymptotic convergence behavior 
compares favorably with other algorithms that
have proven convergence, as illustrated in Fig.~\ref{fig:cost_history}.

One approach to avoid the singularity issue is to use the differentiable penalty function
\begin{equation}
	\label{eq:Reps}
	R_{\epsilon}(\x) = \sum_{i\in\calI}\, \biggl[ \Bigl( \sum_{j\in\calJ}|x(i+j)|^2 \Bigr) + \epsilon \biggr]^{\half}
\end{equation}
where $\epsilon$ is a small positive value, instead of \eqref{eq:ogcost}.
However, as in \cite{Oliveira_2009_SP}, we have found it unnecessary to do so,
since we have found that the algorithm is not hindered by the singularity issue in practice.
For the regularizer \eqref{eq:Reps}, convergence of
the corresponding form of OGS  can be proven 
using the Global Convergence Theorem (GST) \cite{Luenberger_2008, Rao_1999_TSP}.

\smallskip
\noindent
\textbf{Proximal operator.}
An effective approach for solving a wide variety of inverse problems is
given by the proximal framework 
\cite{Combettes_2005, Combettes_2010_chap, Beck_2009_SIAM}.
In this approach, the solution $\x$ to a general inverse problem 
(e.g.\ deconvolution, interpolation, reconstruction) 
with penalty function $R(\x)$
can be computed by solving a sequence of denoising problems
each with penalty function $R(\x)$.
Therefore, the efficient computation of the solution to the denoising
problem is important for the use of proximal methods.
In this context, the denoising problem, i.e.\ \eqref{eq:opt}, is termed the \emph{proximal operator} (or proximity operator).
The OGS algorithm is therefore an implementation of the proximal operator
for penalty function \eqref{eq:ogcost},
and can be used in proximal algorithms for 
inverse problems that are more general than denoising.
As noted above, other implementations of the proximal operator
associated with overlapping group sparsity have been given
\cite{Bach_2012_techrep, Deng_2011_techrep, Jacob_2009_cnf, Bayram_2011_picassp, Boyd_2011_admm, Chen_2012_AAS, Figueiredo_2011_SPARS, Yuan_2011_NIPS, Peyre_2011_eusipco, Jenatton_2010_ICML};
these algorithms are based on the duplication of variables, 
and hence require more memory (proportional to the group size in the fully-overlapping case).

\smallskip
\noindent
\textbf{FOCUSS.}
The OGS algorithm can be considered as a type of FOCUSS algorithm \cite{Gorodnitsky_1997_TSP}
that
is designed to yield sparse solutions to under-determined linear systems of equations.
It was extended to more general penalty functions (or diversity measures) \cite{Rao_1999_TSP}
and to the noisy-data case \cite{Rao_2003_TSP}.
Setting $ p = 1 $ and $ A = \I $ in equation (13) of \cite{Rao_2003_TSP}
gives the OGS algorithm for group size $ K = 1 $,
namely an iterative implementation of the soft-threshold rule.
(Note, the emphasis of FOCUSS is largely on the non-convex ($p<1$) case
with $ A \neq \I $ and only group size $ K = 1 $).

The FOCUSS algorithm was further extended to the case of multiple measurement vectors (MMV)
that share a common sparsity pattern \cite{Cotter_2005_TSP}.
We note that the resulting algorithm,  M-FOCUSS, is different than OGS.
In the MMV model, the location (indices) of significant coefficients is consistent among 
a set of vectors;
while in the OGS problem there is no common (repeated)
sparsity pattern to be exploited.

The M-FOCUSS algorithm was later improved to account for gradual changes in the sparsity pattern 
in a sequence of measurement vectors \cite{Zdunek_2008_TSP}.
This approach involves, in part, arranging the sequence of measurement vectors into overlapping blocks.
While both, the 
algorithm of \cite{Zdunek_2008_TSP} 
and the OGS algorithm,
utilize overlapping blocks, 
the former algorithm utilizes overlapping blocks to
exploit a sparsity pattern (approximately) common to multiple measurement vectors,
whereas the latter does not assume any common sparsity among blocks.

The FOCUSS algorithm was extended to mixed norms in \cite{Kowalski_2009_ACHA}
 to attain structured sparsity without overlapping groups.
This approach is extended in \cite{Kowalski_2009_SIVP, Siedenburg_2011_DAFx, Siedenburg_2013_JAES} where
sliding windows are used to account for overlapping groups.
However, as noted in \cite{Kowalski_2009_SIVP}, this approach does not directly correspond to an optimization problem;
hence it does not constitute an implementation of the proximal operator.

\section{Gaussian noise and OGS}
\label{sec:pnoise}

This section addresses the problem of how to set the regularization
parameter $\lambda$ in \eqref{eq:opt} in a simple and direct way
analogous to the `$3\sigma$ rule'.
The use of the $3\sigma$ rule for soft thresholding,
as illustrated in Sec.~\ref{sec:onedimeg}, is simple to apply
because soft thresholding has a simple explicit form.
However, overlapping group shrinkage
has no explicit form.
Therefore, extending the notion of the `$3\sigma$ rule' to OGS is not straight forward.
The question addressed in the following is:
how should $\lambda$ be chosen so that essentially all the noise is eliminated?
In principle, 
the minimum such $\lambda$ should be used, because a larger value will
only cause further signal distortion.

In order to set $\lambda$ so as to reduce Gaussian noise to 
a desired level, the effect of the OGS algorithm on pure i.i.d.~zero-mean unit-variance Gaussian  noise
is investigated.
We examine first the case of group size $K = 1$, 
because analytic formulas can be obtained in this case 
(OGS being soft thresholding for $K=1$).

\begin{figure}
	\centering	
	\includegraphics{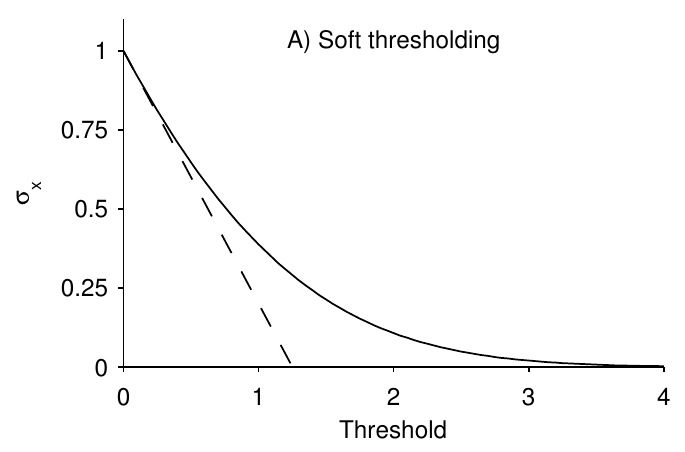}
	\hfill
	\includegraphics{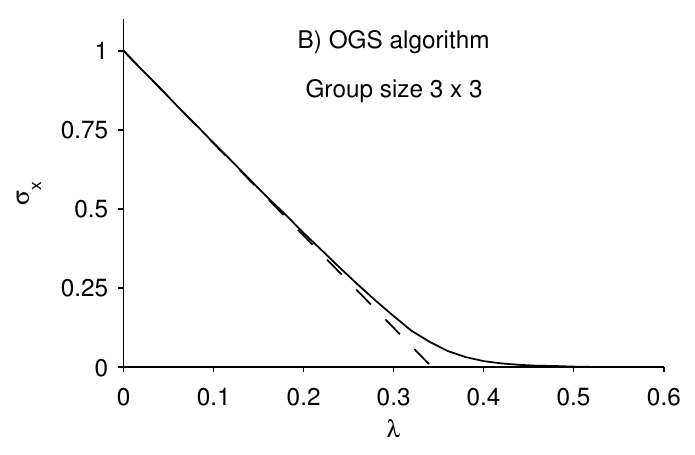}
	\caption{Standard deviation of standard Gaussian noise $\calN(0,1)$ after (a) soft thresholding
	and (b) overlapping group shrinkage (OGS) with group size $ 3 \times 3$.}
	\label{fig:noise_analysis}
\end{figure}

Let $y \sim \calN(0, 1)$ 
and define $ x = \soft(y, \, T)$.
Then the variance of $x$ as a function of threshold $T$ is given by
\begin{align}
	\label{eq:RSV}
	\sigma_x^2(T)
	& = E[x^2]
	= \int_{|y|>T}\, (\abs{y}-T)^2 \, p_y(y) \, dy
	\\
	\label{eq:realsoftvar}
	&
	= 2 \, (1 + T^2) \, Q(T) - T \sqrt{\frac{2}{\pi}} \exp\left(-\frac{T^2}{2} \right) 
\end{align}
where
$ p_y(y) $ is the standard normal pdf $ \calN(0,1)$
and
\[
	Q(T):= \frac{1}{\sqrt{2\,\pi}}\,\int_{T}^{\infty} \myE^{-\frac{t^2}{2}}\,dt
	= 0.5 \,\Bigl(1 - \erf\Bigl(\frac{T}{\sqrt{2}}\Bigr) \Bigr).
\]
The standard deviation $\sigma_x(T)$ is illustrated in Fig.~\ref{fig:noise_analysis}a
as a function of threshold $T$.
Since the variance of $x$ is unity here, the $3\sigma$ rule suggests setting
the threshold to $ T = 3 $ which leads to $\sigma_x(3) = 0.020 $
according to \eqref{eq:realsoftvar}.

The graph in Fig.~\ref{fig:noise_analysis}a generalizes the $3\sigma$ rule:
Given a specified output standard deviation $\sigma_x$, the graph 
shows how to set the threshold $T$ in the soft threshold function so as to achieve it,
i.e., so that $E[\soft(y, T)^2] = \sigma_x^2$ where $y \sim \calN(0, 1)$.
For example, to reduce the noise standard deviation $\sigma$ to 
one percent of its value, we solve $ \sigma_x(T) = 0.01 $ for $T$
to obtain $ T = 3.36 \sigma $, a threshold greater than that suggested by the $3\sigma$ rule.

To set the regularization parameter $\lambda$ in the OGS algorithm,
we suggest that the same procedure can be followed.
However, for OGS there is no explicit formula such as \eqref{eq:realsoftvar} relating $\lambda$ to $\sigma_x$.
Indeed, in the overlapping group case,
neither is it possible to reduce
$E[x^2]$ to a univariate integral as in \eqref{eq:RSV}
due to the coupling among the components of $\y$,
nor is there an explicit formula for $\x$ in terms of $\y$, but only a numerical algorithm.

Although no explicit analog of \eqref{eq:realsoftvar} is available for OGS, 
the functional relationship can be found numerically.
Let $\y$ be i.i.d.\ $\calN(0,1)$
and define $\x$ as the output of the OGS algorithm: $\x = \ogs(\y; \, \lambda, \, K)$.
The output standard deviation $\sigma_x$
can be found by simulation as a function of $\lambda$
for a fixed group size. 
For example, consider applying the OGS algorithm to a two-dimensional array $\y$ using a group size of $ 3 \times 3 $.
For this group size, $\sigma_x$ as a function of $\lambda$ is illustrated in Fig.~\ref{fig:noise_analysis}b.
The graph is obtained by generating a large two-dimensional array
of i.i.d.\ standard normal random variables, applying the OGS algorithm
for a discrete set of $\lambda$, and then computing the standard deviation of the result for each $\lambda$.
Once this graph is numerically obtained,
it provides a straight forward way to set $\lambda$ so as to reduce
the noise to a specified level.
For example, to reduce the noise standard deviation $\sigma$ down to one percent of its value,
we should use $\lambda \approx 0.43 \sigma$ in the OGS algorithm
according to the graph in Fig.~\ref{fig:noise_analysis}b.
(Obtaining the value $\lambda$ corresponding to a specified $\sigma_x$
generally requires an extrapolation between the data points
comprising a graph such as Fig.~\ref{fig:noise_analysis}b.)
Note that the graph will depend on the group size, so
for a different group size the graph (or table) needs to be recalculated.
For efficiency, these calculations can be performed off-line
for a variety of group sizes and stored for later use.

Table~\ref{tab:lamreal} gives the value of the regularization parameter $\lambda$ so
that OGS produces an output signal $\y$ with specified standard deviation $ \sigma_x $
when the input $\y$ is standard normal (zero-mean unit-variance i.i.d.\ Gaussian).
The table applies to both 1D and 2D signals.
The table provides values for groups up to length 5 in 1D,
and up to size $ 5 \times 5 $ in 2D.
Note for groups of size $ 1 \times K $, the value of $\lambda$ is the
same for 1D and 2D.
Also, note that $\lambda$ is the same for groups of size $K_1 \times K_2$ and $K_2 \times K_1$;
so each is listed once in the table.
The first value in each column is obtained using 150 iterations of the OGS algorithm (more than
sufficient for accurate convergence);
while the second value in parenthesis is obtained using 25 iterations (sufficient in practice
and uses less computation).
For group size $ 1 \times 1 $, OGS is simply soft thresholding; hence no iterations
are needed for it.
The values $\lambda$ listed in Table~\ref{tab:lamreal} are accurate to within $0.01$
and were computed via simulation.

To clarify how Table~\ref{tab:lamreal} is intended to be used, suppose one is using the OGS
algorithm with $2 \times 3$ groups for denoising a signal contaminated by
additive white Gaussian noise with standard deviation $\sigma$.
To reduce the noise down to $0.1\%$ of its value, one would use
$\lambda = 0.74 \sigma$ if running the OGS algorithm to full
convergence;
or one would use 
$\lambda = 0.77 \sigma$ if only 25 iterations of the OGS algorithm are used.
The values are from the $10^{-3}$ column in Table \ref{tab:lamreal}.

\begin{table}[t!]
\caption{
	Parameter $\lambda$ for standard normal i.i.d.\ signal
	\label{tab:lamreal}
}
\medskip
\centering
\begin{tabular}{@{}lcccc@{}}
\toprule
 & \multicolumn{4}{c}{Output std $\sigma_x$} \\ 
    \cmidrule(l){2-5}
Group        &   $10^{-2}$ & $10^{-3}$ & $10^{-4}$& $10^{-5}$\\
\midrule
$1 \times 1$     &   $3.36$ & $4.38$ & $5.24$ & $6.00$\\

$1 \times 2$	 &   $1.69\,(1.73)$ & $2.15\,(2.24)$ & $2.38\,(2.61)\,$ & $2.46\,(2.94)$\\
$1 \times 3$	 &   $1.16\,(1.18)$ & $1.46\,(1.52)$ & $1.60\,(1.77)$ & $1.64\,(1.99)$\\
$1 \times 4$	 &   $0.89\,(0.91)$ & $1.12\,(1.16)$ & $1.23\,(1.36)$ & $1.27\,(1.53)$\\
$1 \times 5$	 &   $0.73\,(0.75)$ & $0.92\,(0.95)$ & $1.01\,(1.12)$ & $1.04\,(1.25)$\\
$2 \times 2$     &   $0.86\,(0.87)$ & $1.08\,(1.13)$ & $1.19\,(1.31)$ & $1.23\,(1.48)$ \\ 
$2 \times 3$	 &   $0.59\,(0.61)$ & $0.74\,(0.77)$ & $0.80\,(0.89)$ & $0.82\,(1.01)$\\
$2 \times 4$     &   $0.46\,(0.48)$ & $0.57\,(0.59)$ & $0.62\,(0.69)$ & $0.64\,(0.78)$\\
$2 \times 5$	 &   $0.38\,(0.41)$ & $0.46\,(0.49)$ & $0.51\,(0.57)$ & $0.52\,(0.64)$\\
$3 \times 3$	 &   $0.41\,(0.43)$ & $0.50\,(0.53)$ & $0.55\,(0.61)$ & $0.56\,(0.69)$\\
$3 \times 4$  	 &   $0.33\,(0.35)$ & $0.39\,(0.42)$ & $0.43\,(0.48)$ & $0.44\,(0.54)$\\
$3 \times 5$	 &   $0.29\,(0.31)$ & $0.32\,(0.36)$ & $0.35\,(0.40)$ & $0.36\,(0.45)$\\
$4 \times 4$	 &   $0.27\,(0.30)$ & $0.30\,(0.34)$ & $0.33\,(0.38)$ & $0.34\,(0.43)$\\
$4 \times 5$	 &   $0.24\,(0.26)$ & $0.26\,(0.30)$ & $0.27\,(0.33)$ & $0.28\,(0.37)$\\
$5 \times 5$	 &   $0.21\,(0.23)$ & $0.22\,(0.26)$ & $0.23\,(0.29)$ & $0.24\,(0.32)$\\
$2 \times 8$     &   $0.28\,(0.30)$ & $0.31\,(0.35)$ & $0.33\,(0.39)$ & $0.35\,(0.43)$\\ 
\bottomrule
\\
\multicolumn{5}{@{}l}{
	Regularization parameter $\lambda$ to achieve specified output standard}
\\
\multicolumn{5}{@{}l}{
	deviation when OGS is applied to a real standard normal signal:}
\\
\multicolumn{5}{@{}l}{
	full convergence (25 iterations).}
\end{tabular}

\end{table}

It can be observed in Fig.~\ref{fig:noise_analysis} that 
the function $\sigma_x(\cdot)$ has a sharper `knee'
in the case of OGS compared with soft thresholding.
We have examined graphs for numerous group sizes
and found that in general the larger the group, the sharper is the knee.
Note that in practice $\lambda$ should be chosen 
large enough to reduce the noise to a sufficiently negligible level,
yet no larger so as to avoid unnecessary signal distortion.
That is, suitable values of $\lambda$ are somewhat near the knee.
Therefore, due to the sharper knee,
the denoising process is more sensitive to $\lambda$
for larger group sizes; hence, the choice of $\lambda$ is more critical.

Similarly, it can be observed in Fig.~\ref{fig:noise_analysis} that
for OGS, the function $\sigma_x(\cdot)$ follows
a linear approximation more closely to the left of the `knee'
than it does in the case of soft thresholding.
We have found that near the origin,
$\sigma_x(\lambda)$ is approximated by
\begin{equation}
	\label{eq:slope}
	\sigma_x(\lambda) \approx -\sqrt{2} \, \frac{\Gamma(\abs{\calJ}/2+1/2)}{\Gamma(\abs{\calJ}/2)} \, \lambda,
	\quad
	\text{for} \; \lambda \approx 0,
\end{equation}
where $\abs{\calJ}$ is the cardinality of the group ($K$ in 1D, $K_1 K_2$ in 2D).
This can be explained by noting that for $\y \sim \calN(0,\sigma^2)$,
the $\ell_2$ norm of the group follows a chi-distribution with $\abs{\calJ}$ degrees of freedom,
the mean of which is the slope in \eqref{eq:slope}.
For small $\lambda$, OGS has roughly the effect of soft thresholding 
the $\ell_2$ norm of the groups.
However, it should be noted that small $\lambda$ are of minor interest 
in practice because in this range, noise is not sufficiently attenuated.
The right hand side of \eqref{eq:slope} is illustrated as dashed lines in Fig.~\ref{fig:noise_analysis}.

\subsection{Shrinkage/Thresholding behavior}

\begin{figure}[t!]
	\centering	
	\includegraphics{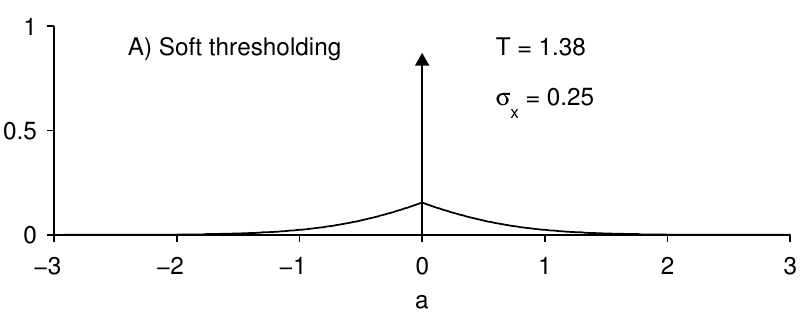} \\[1em]
	\includegraphics{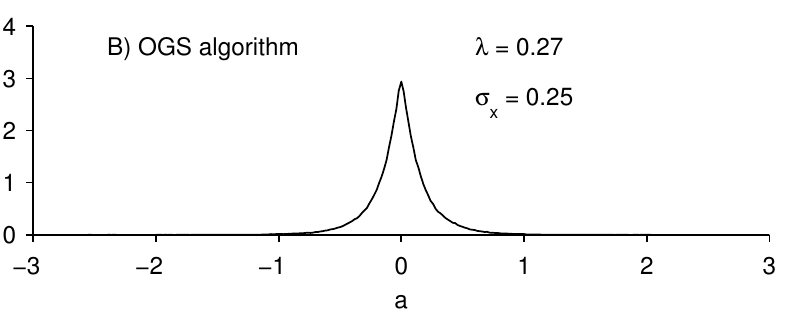} \\[1em]
	\includegraphics{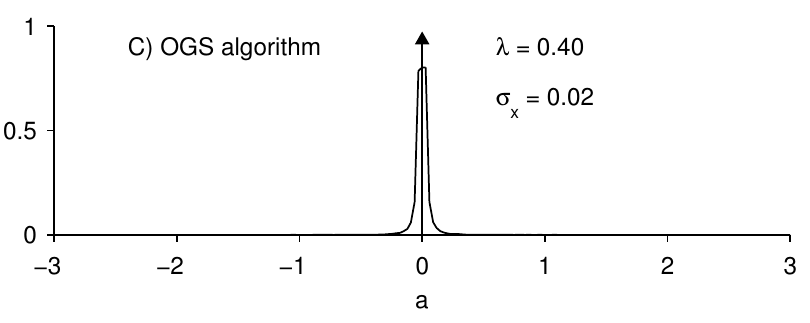}
	\caption{Probability density function of zero-mean unit-variance Gaussian noise
	after soft thresholding (a) and OGS (b,c).
	In (a) and (b), the parameters ($T$ and $\lambda$) are set so that $\sigma_x = 0.25$.
	In (b) and (c) OGS was applied with group size $3 \times 3$.
	In (a) and (c) the pdf contains a point mass at zero.
	}
	\label{fig:noise_analysis_pdf}
\end{figure}

Although not apparent in Fig.~\ref{fig:noise_analysis}, the soft-thresholding
and OGS procedures are quite distinct in their shrinkage/thresholding behavior.
Clearly, if $\y$ is i.i.d.~with $ y \sim \calN(0,1)$ and if $\x$ is the result of 
soft thresholding (i.e.\ $ x = \soft(y, T)$), then $\x$ contains many zeros.
All values $ \abs{y} \le T $ are mapped to zero.
The nonlinearity involves \emph{thresholding} in this sense.
In contrast, OGS does not produce any zeros unless $\lambda$ is
sufficiently large.
This behavior is illustrated in Fig.~\ref{fig:noise_analysis_pdf},
which shows the pdf of $y$ after soft thresholding (a) and after OGS (b, c).
Soft thresholding with $ T = 1.38 $ produces an output $x$ with $\sigma_x = 0.25 $.
The pdf of $x$, illustrated in Fig.~\ref{fig:noise_analysis_pdf}a,
consists of a point mass at the origin of mass $0.831$
and the tails of the Gaussian pdf translated toward the origin.
The point mass represents the zero elements of $x$.

Using OGS with group size $ 3 \times 3 $ and 
$\lambda$ set so that again $\sigma_x = 0.25 $,
the output $x$ does not contain 
zeros.
The pdf is illustrated in Fig.~\ref{fig:noise_analysis_pdf}b.
The absence of the point mass at the origin reflects
the fact that OGS mapped no values of $y$ to zero,
i.e. no thresholding.
The pdf in Fig.~\ref{fig:noise_analysis_pdf}b is computed
numerically by applying the OGS algorithm to simulated standard normal data,
as no explicit formula is available.

When $\lambda$ is sufficiently large, then OGS does perform thresholding,
as illustrated in Fig.~\ref{fig:noise_analysis_pdf}c.
For a group size of $ 3 \times 3 $ and $ \lambda = 0.4$, the pdf
exhibits a point-mass at the origin
reflecting the many zeros in the output of OGS.\footnote{The pdf in Fig.~\ref{fig:noise_analysis_pdf}c is computed as a histogram; hence, the exact value of the point-mass at the origin depends on the histogram bin width.} 
For this value of $\lambda$, OGS performs both thresholding and shrinkage, like the soft threshold function.

\begin{table}[t!]
\caption{
	Parameter $\lambda$ for standard complex normal i.i.d.\ signal
	\label{tab:lamcomplex}
}
\medskip
\centering
\begin{tabular}{@{}lcccc@{}}
\toprule
 & \multicolumn{4}{c}{Output std $\sigma_x$} \\ 
    \cmidrule(l){2-5}
Group        &   $10^{-2}$ & $10^{-3}$ & $10^{-4}$& $10^{-5}$\\
\midrule
$1 \times 1$     &   $2.54$ & $3.26$ & $3.86$ & $4.39$\\

$1 \times 2$	 &   $1.30\,(1.33)$ & $1.65\,(1.71)$ & $1.82\,(2.00)$ & $1.89\,(2.25)$\\
 
$1 \times 3$	 &   $0.90\,(0.93)$ & $1.12\,(1.17)$ & $1.22\,(1.35)$ & $1.26\,(1.52)$\\

$1 \times 4$	 &   $0.71\,(0.73)$ & $0.86\,(0.91)$ & $0.93\,(1.04)$ & $0.96\,(1.17)$\\

$1 \times 5$	 &   $0.60\,(0.62)$ & $0.71\,(0.75)$ & $0.76\,(0.86)$ & $0.78\,(0.96)$\\

$2 \times 2$     &   $0.66\,(0.69)$ & $0.83\,(0.87)$ & $0.91\,(1.01)$ & $0.94\,(1.13)$\\ 

$2 \times 3$	 &   $0.48\,(0.51)$ & $0.56\,(0.60)$ & $0.61\,(0.69)$ & $0.63\,(0.78)$\\
 
$2 \times 4$     &   $0.39\,(0.42)$ & $0.44\,(0.49)$ & $0.47\,(0.55)$ & $0.49\,(0.61)$\\
 
$2 \times 5$	 &   $0.34\,(0.37)$ & $0.37\,(0.42)$ & $0.39\,(0.47)$ & $0.41\,(0.53)$\\

$3 \times 3$	 &   $0.36\,(0.39)$ & $0.40\,(0.45)$ & $0.42\,(0.50)$ & $0.43\,(0.56)$\\

$3 \times 4$  	 &   $0.30\,(0.33)$ & $0.32\,(0.38)$ & $0.34\,(0.42)$ & $0.35\,(0.47)$\\

$3 \times 5$	 &   $0.27\,(0.29)$ & $0.28\,(0.33)$ & $0.29\,(0.37)$ & $0.30\,(0.41)$\\

$4 \times 4$	 &   $0.26\,(0.28)$ & $0.27\,(0.32)$ & $0.28\,(0.36)$ & $0.29\,(0.40)$\\

$4 \times 5$	 &   $0.23\,(0.25)$ & $0.24\,(0.28)$ & $0.25\,(0.32)$ & $0.25\,(0.35)$\\

$5 \times 5$	 &   $0.20\,(0.22)$ & $0.21\,(0.25)$ & $0.22\,(0.28)$ & $0.22\,(0.31)$\\

$2 \times 8$     &   $0.26\,(0.28)$ & $0.27\,(0.32)$ & $0.28\,(0.36)$ & $0.29\,(0.40)$\\ 
\bottomrule
\end{tabular}

\end{table}

\subsection{Complex data}

The calculation \eqref{eq:realsoftvar} is for real-valued standard normal $x$.  
For complex-valued Gaussian data the formula is slightly different:
\begin{align}
	\sigma_x^2	& 	= \int_{\abs{y}>T}\, \abs{\abs{y}-T}^2 \, p_y(y) \, dy	\\
	& = \exp(-T^2) - 2\sqrt{\pi} \, T \,Q(\sqrt{2} \,T)
	\label{eqn: var}  	
\end{align}
where $p_y(y)$ is the zero-mean unit-variance complex Gaussian pdf (standard complex normal), $\mathcal{CN}(0,1)$.
In the complex case, \eqref{eq:slope} is modified to
\begin{equation}
	\label{eq:slope_cplx}
	\sigma_x(\lambda) \approx - \frac{\Gamma(\abs{\calJ}+1/2)}{\Gamma(\abs{\calJ})} \, \lambda,
	\quad
	\text{for} \; \lambda \approx 0,
\end{equation}
as the degrees of freedom of the chi-distribution is doubled (due to real and imaginary parts
of complex $y$).
Because complex-valued data is common (using the Fourier transform, STFT,
and in radar and medical imaging, etc.), it is useful to address the selection
of $\lambda$ in the complex case as well as in the real case.
Note that Table~\ref{tab:lamreal} does not apply to the complex case.
Table~\ref{tab:lamcomplex} gives the value of $\lambda$ for the complex case,
analogous to  Table~\ref{tab:lamreal} for the real case.

\subsection{Remarks}

The preceding sections described
how the parameter $\lambda$ may be chosen
so as to reduce additive white Gaussian noise to a desired level.
However, in many cases the noise is not white.
For example, in the speech denoising example in Section~\ref{sec:speech}, the OGS algorithm
is applied directly in the STFT.
However, the STFT is an overcomplete transform;
therefore, the noise in the STFT domain will not be white, even if it is white in the original signal domain.
In the speech denoising example in Section~\ref{sec:speech} below, this issue is ignored and the algorithm is still effective.
However, in other cases, where the noise is more highly correlated, the values $\lambda$
in Table~\ref{tab:lamreal} and \ref{tab:lamcomplex} may be somewhat inaccurate.

The penalty function \eqref{eq:ogcost} is suitable for stationary noise; however, in many
applications,  noise is not stationary.
For example, in the problem of denoising speech corrupted by stationary colored noise,
the variance of the noise in the STFT domain will vary as a function of frequency.
In particular, some noise components may be narrowband and therefore
occupy a narrow time-frequency region. 
The OGS penalty function and algorithm, as described in this paper, do not apply to this problem directly.
The penalty function \eqref{eq:ogcost} 
and the process to select $\lambda$ 
must be appropriately modified.

The OGS algorithm as described above uses the same block size over the entire signal.
In some applications, it may be more appropriate that the block size varies.
For example, in speech denoising, as noted and developed in \cite{Yu_2008_TSP},
it is beneficial that the block size in the STFT domain varies as a function of frequency (e.g.\ for higher temporal 
resolution at higher frequency).
In addition, the generalization of OGS so as to denoise wavelet coefficients on tree-structured models
as in \cite{Rao_2011_ICIP} may be of interest.

\section{Examples}
\label{sec:examples}

\subsection{One-dimensional denoising}
\label{sec:onedimeg}

\begin{figure}[t!]
	\centering
	\includegraphics{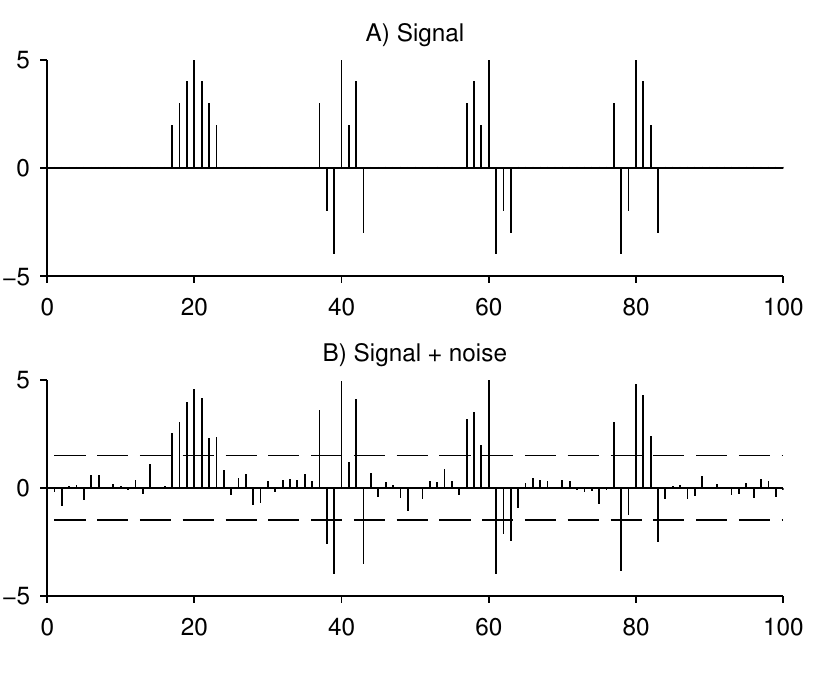}
	
	\includegraphics{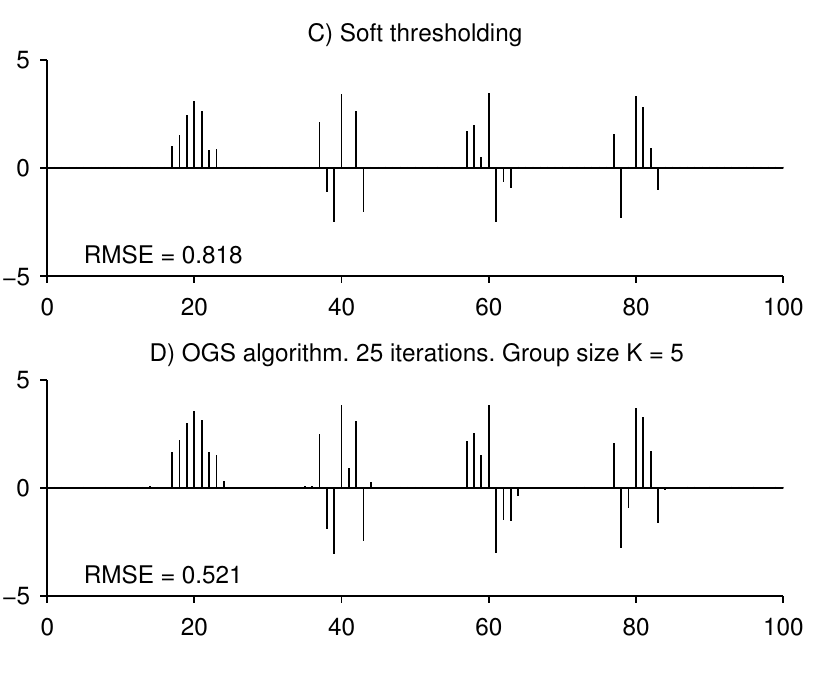}
	
	\includegraphics{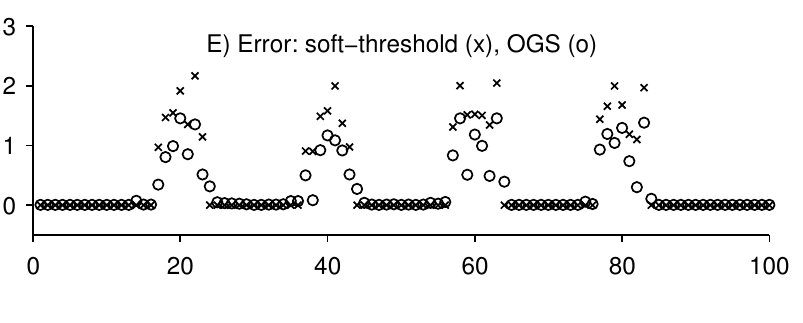}	

	\caption{Group sparse signal denoising: comparison of soft thresholding and 
	overlapping group shrinkage (OGS). OGS yields the smaller RMSE.}
	\label{fig:example1}
\end{figure}

As an illustrative example, the OGS algorithm is applied to denoising 
the one-dimensional group sparse signal in Fig.~\ref{fig:example1}a.
The noisy signal in Fig.~\ref{fig:example1}b is obtained by
adding independent white Gaussian noise with standard deviation $\sigma = 0.5$.
The dashed line in the figure indicates the `$3\sigma$' level.
The `$3\sigma$ rule' states that nearly all values of a Gaussian 
random variable lie within three standard deviations
of the mean (in fact, $99.7\%$).
Hence, by using $3\sigma$ as a threshold with the soft threshold function,
nearly all the noise will be eliminated as illustrated in 
Figure~\ref{fig:example1}c 
with threshold $ T = 3\sigma = 1.5 $.
Although the noise is effectively eliminated, the large-amplitude values of the signal 
have been attenuated, which is unavoidable when applying soft thresholding.

Even though this choice of threshold value does not minimize the RMSE,
it is a simple and intuitive procedure which
can be effective and practical in certain applications.
Moreover, this rule does not depend on the signal energy (only the noise variance), so it is 
straight-forward to apply.
Regarding the RMSE, it should be noted that optimizing the RMSE 
does not always lead to the most favorable denoising result in practice.
For example, in speech enhancement/denoising,
even a small amount of residual noise will be audible as `musical noise' \cite{Loizou_2007}.
(In speech denoising via STFT thresholding, the RMSE-optimal threshold 
produces a denoised signal of low perceptual quality.)

\begin{figure}[t!]
	\centering
	\includegraphics{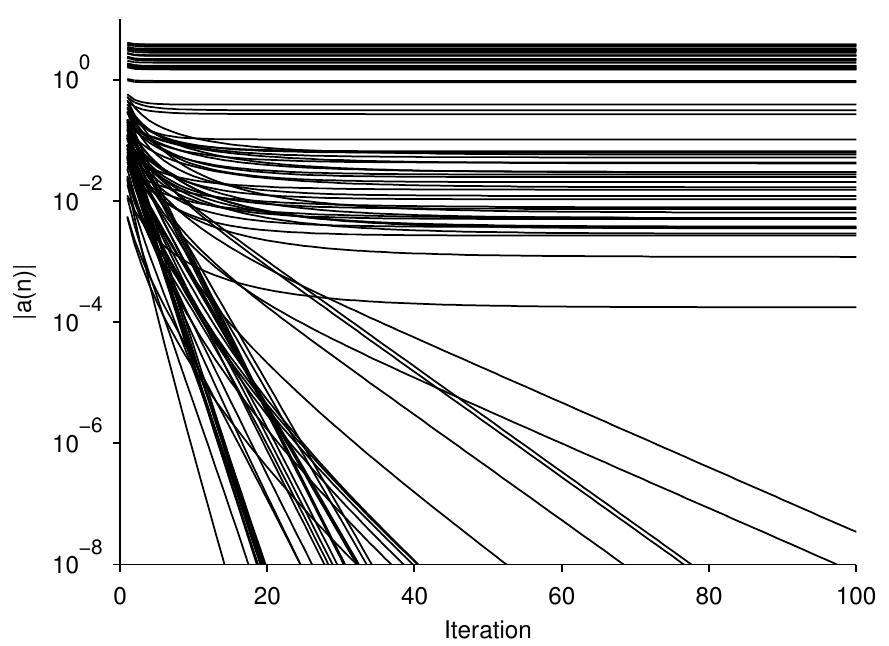}
	\caption{Convergence of OGS algorithm as applied in Fig.~\ref{fig:example1} .}
	\label{fig:example1_conv}
\end{figure}

The result of applying the OGS algorithm to the noisy signal
is illustrated in Fig.~\ref{fig:example1}d.
Twenty-five iterations of the algorithm were used,  with group size $K = 5 $
and parameter\footnote{The parameter $\lambda$ is chosen so that 
25 iterations of the OGS algorithm with group size $K=5$ applied to white Gaussian noise produces
the same output standard deviation as soft thresholding using threshold $T = 3 \sigma$.
This method for selecting $\lambda$ is elaborated in Section \ref{sec:pnoise}.} $ \lambda = 0.68 \sigma = 0.34 $.
As visible, the noise has been essentially eliminated.
Compared to soft thresholding in Fig.~\ref{fig:example1}c, 
the large-amplitude values of the original signal are less attenuated,
and hence the RMSE is improved ($0.52$ compared to $0.82$).
In this example, the RMSE comparison between the two methods
is meaningful because both methods have been used with
parameter values chosen so as to eliminate essentially all the noise.

The convergence of $ \x^{(k)} $ to $ \x\opt $ is illustrated in Fig.~\ref{fig:example1_conv},
where it can be seen that numerous values $x^{(k)}(i)$ converge to zero.
In this example, with initialization $ x^{(0)}(i) \neq 0, \, \forall i \in \calI $,
we have  $ x^{(k)}(i) \neq 0, \, \forall i \in \calI $ for all subsequent iterations $ k \ge 1 $.
Yet, entire groups converge to zero.

\begin{figure}
	\centering
	\includegraphics{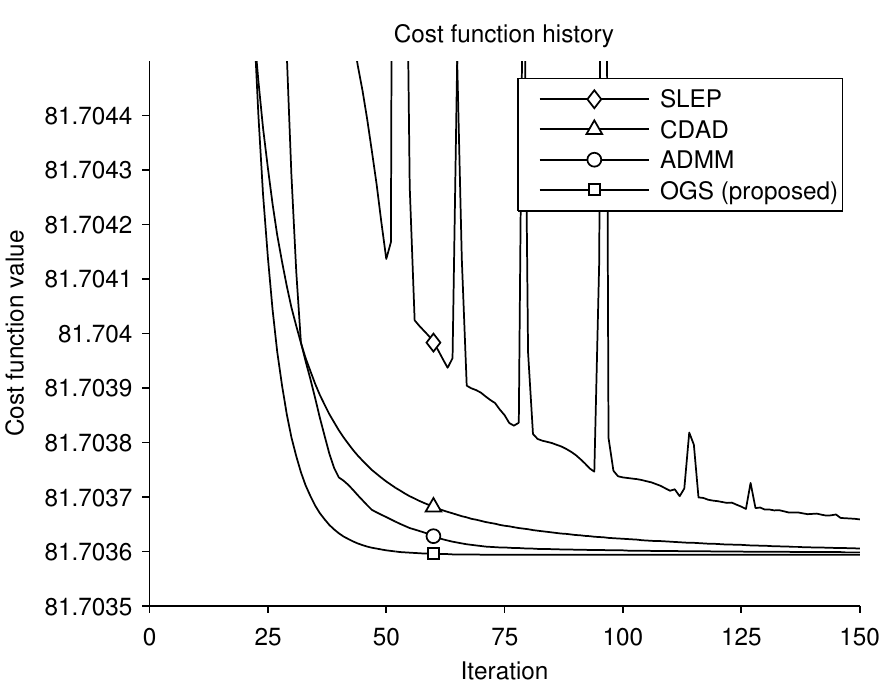}
	\caption{
		Cost function history: comparison of algorithms for 1-D denoising.
	}
	\label{fig:cost_history}
\end{figure}

The convergence behavior in terms of the cost function history of the OGS algorithm is compared 
with three other algorithms in Fig.~\ref{fig:cost_history},
namely to 
ADMM \cite{Deng_2011_techrep},
CDAD (coordinate descent algorithm for the dual problem) \cite{Bayram_2011_picassp, Jenatton_2010_ICML},
and
the algorithm of \cite{Yuan_2011_NIPS}  as implemented in the  SLEP (Sparse Learning with Efficient Projections) software package\footnote{The SLEP software was downloaded from \url{http://www.public.asu.edu/~jye02/Software/SLEP/}. We thank the author for verifying the correctness of our usage of SLEP. We also note that SLEP is a versatile suite of algorithms that can solve a variety of sparsity-related optimization problems.}.
The figure shows the cost function history for the first 150 iterations
of each algorithm.
The OGS algorithm exhibits a monotone decreasing cost function
that attains the lowest cost function value after 25 iterations.
The ADMM, CDAD, and OGS 
algorithms have about the same computational cost per iteration, $O(NK)$.
We implemented these three algorithms in MATLAB.
The SLEP software is written in C (compiled in MATLAB as a mex file)
and 
each iteration of SLEP performs several inner optimization iterations.

The run-time for 150 iterations of SLEP, CDAD, ADMM, and OGS are 3, 99.5, 36.7 and 7.7  milliseconds, respectively.
Note that SLEP and OGS are much faster than the other two.
SLEP is written in C/Mex, so it is difficult to compare its run-time with the other three algorithms
which are written in MATLAB.
The MATLAB implementation of OGS is fast due to
\ia\ its minimal data indexing,
\ib\ its minimal memory usage (no auxiliary/splitting variables),
\ic\ its computational work is dominated by convolution which is implemented very efficiently with the MATLAB built-in \texttt{conv} function.

It should also be noted that ADMM requires that parameters be specified by the user,
which we manually set  so as to optimize its convergence behavior.
The algorithm of \cite{Yuan_2011_NIPS} also calls for parameters,
but the SLEP software provides default values which we used here.
The CDAD and OGS algorithms do not require any user
parameters to be specified.
We also note that ADMM and CDAD require 5-times the memory of OGS,
as the group size is $K = 5$ and the groups are fully overlapping.
In summary, the OGS algorithm has minimal memory requirements,
requires no user specified parameters, and has the most favorable asymptotic convergence
behavior.

We have found that for some problems 
OGS has a slow convergence during the initial iterations,
compared to the other algorithms.
This is because OGS does not perform explicit thresholding as do the other
algorithms; instead, OGS gradually reduces values toward zero.
It may be appropriate to perform several iterations of another
algorithm or preliminary thresholding 
to improve initial convergence, and then switch to OGS.
The comparison here uses OGS alone.

\medskip

\noindent
\textbf{Partially overlapping groups.}
It may be expected that partially overlapping groups may also serve as a useful signal model.
To this end, we performed numerical experiments to evaluate the utility of partially overlapping
group sparsity by modifying the penalty function \eqref{eq:ogcost} so that the outer
sum is over a sub-grid of $\calI$. 
We used a set of signals like that of Fig.~\ref{fig:example1}a where each group is systematically
translated in turn, and we modified the OGS algorithm accordingly.
For each signal, the RMSE-optimal $\lambda$ was numerically found, and the corresponding
optimal RMSE recorded for several values of the noise variance $\sigma^2$.
Averaging over 100 realizations for each group position and overlap, we obtain the RMSE
values shown in Table~\ref{tab:rmse_overlap}.
The fully-overlapping case (overlap $M=4$) gives the lowest RMSE, as might be expected.
However, the non-overlapping case (overlap $M=0$) does not give the worst RMSE.
It turns out that partial overlapping can yield an inferior RMSE compared to 
both fully-overlapping and non-overlapping cases.
The reason for this is that, in the partial overlapping case, some components of $\x$ will 
be a member of only one group, while other components will be a member of two or more groups.
The more groups a component is a member of, the more strongly it is penalized.
Hence, components of $\x$ are non-uniformly penalized in the partially-overlapping case,
and this degrades its performance when the group structure is not known \textit{a priori}.

\begin{table}
\label{tab:rmse_overlap}
\centering
\caption{Average minimum RMSE for partial overlap ($K = 5$)}
\medskip
\begin{tabular}{@{}lccccc@{}}
\toprule
		&  \multicolumn{5}{c}{overlap $M$} \\ \cmidrule(l){2-6}
    $\sigma$ &  0	    &  1	   &	2	  &	3	     &	4\\ \midrule
	0.5			 &	0.3925  &  0.3978  &  0.3935  &   0.3886  &  0.3846	\\
	1	         &	0.7282  &  0.7391  &  0.7344  &   0.7248  &  0.7176	\\
	2	         &	1.2135  &  1.2303  &  1.2291  &   1.2080  &  1.1954	\\
\bottomrule
\end{tabular}
\end{table}

\subsection{Speech Denoising}
\label{sec:speech}

This section illustrates the application of overlapping group shrinkage (OGS) to the problem
of speech enhancement (denoising) \cite{Loizou_2007}.
The short-time Fourier transform (STFT) is the basis of many algorithms for 
speech enhancement, including 
classical spectrum subtraction \cite{Boll_1979,McAulay_1980},
improved variations thereof using non-Gaussian models \cite{Martin_2005_TSAP,Hendriks_2007_taslp},
and methods based on Markov models \cite{Dikmen_2010_TASLP, Fevotte_2008_TSALP, Wolfe_2004_royal}.
A well known problem arising in many speech enhancement algorithms is
that the residual noise is 
audible as `musical noise' \cite{Berouti_1979}.
Musical noise may be attributed to isolated noise peaks in the time-frequency domain that
remain after processing.
Methods to reduce musical noise 
include over estimating the noise variance, imposing a minimum spectral noise floor \cite{Berouti_1979},
and 
improving the estimation of model parameters \cite{Cappe_1994}.
Due to the musical noise phenomenon, it is desirable to reduce the noise sufficiently,
even if doing so is not optimal with respect to the RMSE.

To avoid isolated spurious time-frequency noise spikes (to avoid musical noise),
the grouping/clustering behavior of STFT coefficients of speech waveforms 
can be taken into account.
To this end, a recent algorithm by Yu et al.\ \cite{Yu_2008_TSP} for speech/audio enhancement
consists of  time-frequency block thresholding.
We note that the algorithm of \cite{Yu_2008_TSP} is based on non-overlapping blocks.
Like the algorithm of \cite{Yu_2008_TSP}, the OGS algorithm aims to draw on the grouping
behavior of STFT coefficients so as to improve the overall denoising result,
but it uses a model based on fully overlapping blocks.
Some other recent algorithms exploiting structured time-frequency sparsity of speech/audio
are based on Markov models \cite{Dikmen_2010_TASLP, Fevotte_2008_TSALP, Wolfe_2004_royal}.

\begin{figure}[t!]
	\begin{center}	
	\begin{tabular}{cc}
		\includegraphics{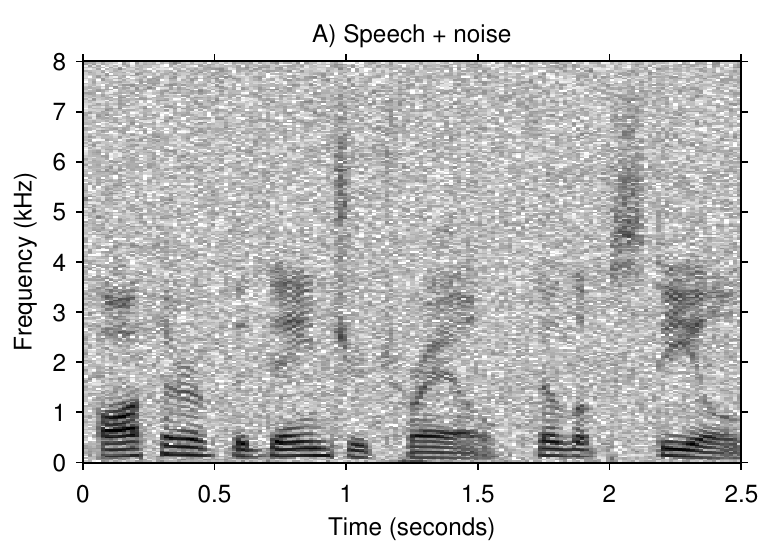}	
		&
		\includegraphics{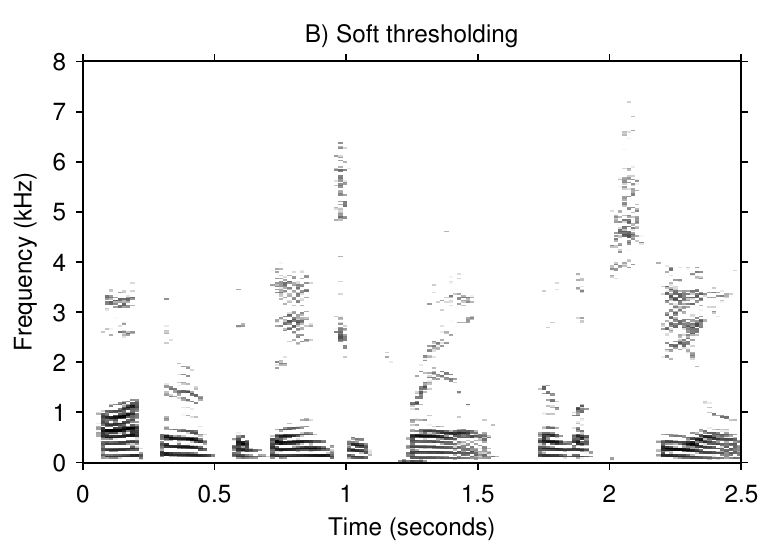}
		\\
		\includegraphics{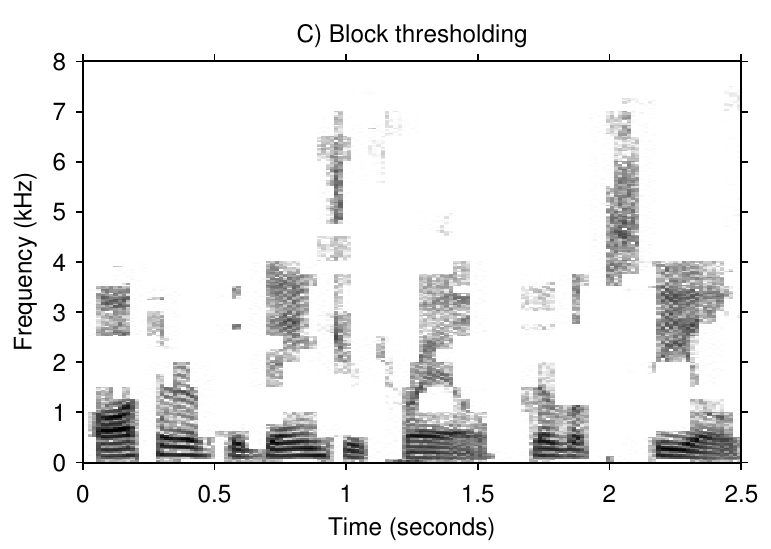} 
		&
		\includegraphics{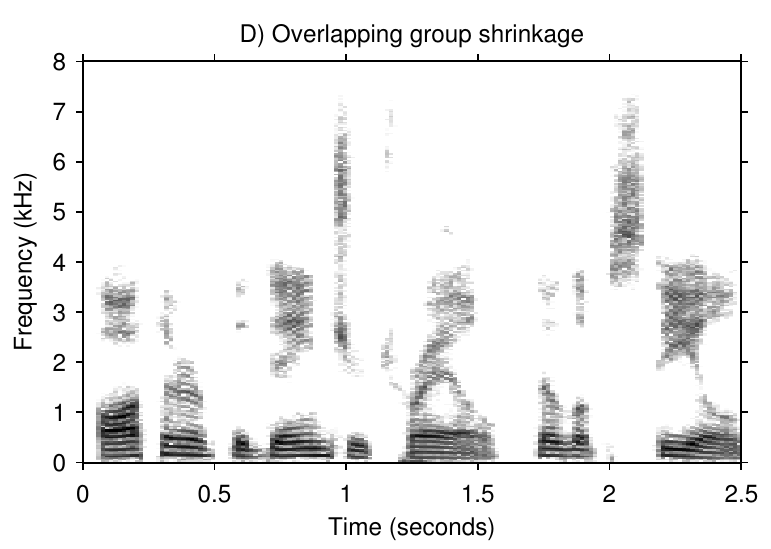}  
	\end{tabular}
	\end{center}
	\caption{
		Spectrograms of (a) noisy speech and (b) result of soft thresholding.
		Spectrograms denoised by  (c) block thresholding \cite{Yu_2008_TSP} and (d) overlapping group shrinkage (OGS).
		Gray scale represents decibels.}
	\label{fig:speech}
\end{figure}

To apply OGS to speech denoising, we solve \eqref{eq:opt},
where
$\y = \Phi \s $ is the complex short-time Fourier transform (STFT) of the noisy speech $\s$,
$\Phi$ is the STFT,
$\x$ is the STFT coefficients to be determined,
and $R(\x)$ is \eqref{eq:ogcost}.
The estimated speech is then given by $\hat{\x} = \Phi\ct \x $.
To minimize \eqref{eq:opt}, we use the two-dimensional form of the OGS algorithm applied to the STFT coefficients.

For illustration, consider the noisy speech illustrated in Fig.~\ref{fig:speech}a.
The noisy speech is a male utterance sampled at 16 kHz,
corrupted by additive white Gaussian noise with SNR 10~dB.\footnote{Speech file  \url{arctic_a0001.wav} downloaded from \url{http://www.speech.cs.cmu.edu/cmu_arctic/cmu_us_bdl_arctic/wav}. }
The STFT is calculated with 50\% overlapping blocks of length of 512 samples.
Figure \ref{fig:speech}b illustrates the STFT obtained by soft thresholding the noisy STFT,
with threshold $T$ selected
so as to reduce the noise standard deviation down to $0.1\%$ of its value.
From Table~\ref{tab:lamcomplex}, we obtain $ T = 3.26 \sigma $,
where $\sigma$ is the noise standard deviation in the STFT domain.
The noise is sufficiently suppressed so that musical noise is not audible;
however, the signal is somewhat distorted due to the relatively high 
threshold used.
The spectrogram in Fig.~\ref{fig:speech}b is perhaps overly thinned.

Figure~\ref{fig:speech}c illustrates the result of block thresholding \cite{Yu_2008_TSP}
using the software provided by the authors.\footnote{\url{http://www.cmap.polytechnique.fr/~yu/research/ABT/samples.html}} The SNR is 15.35 dB.
It can be seen that BT produces blocking artifacts in the spectrogram.
Some auditory artifacts are perceptible in the BT denoised speech.

Figure~\ref{fig:speech}d illustrates the result of overlapping group shrinkage (OGS)
applied to the noisy STFT.
We used 25 iterations of the OGS algorithm.
Based on listening to speech signals denoised with various group sizes, we selected a group size $ 8 \times 2 $
(i.e.\ eight frequency bins $\times$ two time bins).
Other group sizes may be more appropriate for other sampling rates and STFT block lengths.
As in the soft thresholding experiment, 
the parameter $\lambda$ was selected 
so as to reduce the noise standard deviation down to $0.1\%$ of its value.
From Table~\ref{tab:lamcomplex}, we obtain $ \lambda = 0.32 \sigma $.
The SNR of the denoised speech is $13.77$ dB.
While the SNR is lower than block thresholding, the artifacts are less audible.
As for soft-thresholding with $\lambda$ selected likewise, musical noise is not audible in the resulting waveform
obtained by applying the inverse STFT.

It was found in \cite{Yu_2008_TSP} that empirical Wiener post-processing (EWP), introduced in \cite{Ghael_1997_SPIE}, 
improves the result of the block thresholding (BT) algorithm.
This post-processing, which is computationally very simple, improves the result of OGS
by an even greater degree than for BT, as measured by SNR improvement.
The Wiener post-processing raises the SNR for BT from 15.35 dB to 15.75 dB,
while it raises the SNR for OGS from 13.77 dB to 15.63 dB.
Hence, the two methods give almost the same SNR after Wiener post-processing.
The substantial SNR improvement in the case of OGS can be explained as follows:
The OGS algorithm has the effect of slightly shrinking (attenuating) large coefficients which produces a bias
and negatively affects the SNR of the denoised signal.
The Wiener post-processing procedure largely corrects that bias; it has the
effect of rescaling (slightly amplifying) the large coefficients appropriately.

\medskip
\noindent
\textbf{Signal-domain data fidelity.}
Note that, due to the STFT not being an orthonormal transform, 
the noise in the STFT domain is not white, even when it is white in the signal domain.
Therefore, instead of problem \eqref{eq:opt}, it is reasonable to solve
\begin{equation}
	\label{eq:optc}
	\min_{\x} \; \frac{1}{2} \, \norm{ \s - \Phi\ct \x }_2^2 + \lambda \, R(\x),
\end{equation}
which is consistent with the white noise assumption (the data fidelity term is in the signal domain).
Problem \eqref{eq:optc} can be solved by 
proximal methods (e.g.~forward-backward splitting (FBS)) \cite{Combettes_2005, Combettes_2010_chap, Beck_2009_SIAM}
or ADMM methods \cite{Deng_2011_techrep, Figueiredo_2011_SPARS, Boyd_2011_admm}.

We note that the solving \eqref{eq:optc} takes more computation than solving \eqref{eq:opt}.
Formulation \eqref{eq:opt} is, by definition, a single proximal operator (which OGS implements).
On the other hand, 
solving \eqref{eq:optc} (using ADMM or FBS, etc.) requires the repeated application of the proximal operator,
interlaced with $\Phi$ and $\Phi\ct$ (forward and inverse STFTs).
In our experiment, solving (34) using ADMM took 5.95 seconds, while solving (2) took 0.25 seconds.
We ran 25 iterations of OGS in each ADMM iteration to implement the proximal operator.

\begin{figure}[t]
	\centering	
	\includegraphics{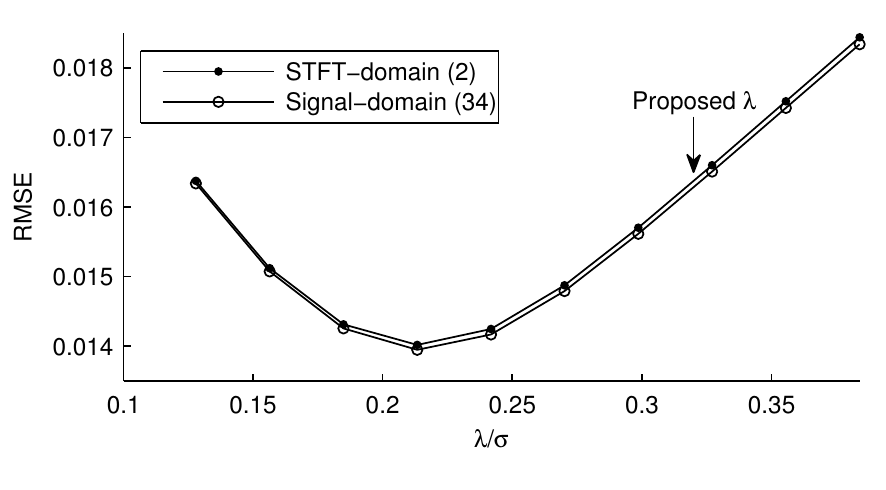}
	\caption{
		Comparison of RMSE using STFT-domain cost function \eqref{eq:opt} 
		and
		signal-domain cost function \eqref{eq:optc}.
		Formulation \eqref{eq:optc} gives a slightly better RMSE.
		The proposed value ($\lambda = 0.32 \sigma $) is indicated by the arrow.
		}
	\label{fig:time_vs_stft}
\end{figure}

Figure \ref{fig:time_vs_stft} shows the RMSE between the estimated speech $\hat{\x}$ and the original speech 
as a function of $\lambda$ for each of \eqref{eq:opt} and \eqref{eq:optc}.
Note that \eqref{eq:optc} attains a better RMSE (higher SNR) than \eqref{eq:opt},
but the difference is marginal and imperceptible. 
We also note that, for both formulations \eqref{eq:opt} and \eqref{eq:optc}, when
$ \lambda $ is optimized so as to obtain the minimum RMSE (best SNR), 
`musical noise' is clearly audible in the denoised speech.
In terms of perceptual quality, a higher value of $ \lambda $ gives a much better result.
Our proposed method for setting $\lambda$, indicated by the arrow in Fig.~\ref{fig:time_vs_stft}, 
gives a higher RMSE, but the result is perceptually preferable
due to the suppression of `musical noise'.
Note that the increase in RMSE, due to using a higher $\lambda$ so as to
avoid musical noise, outweighs the increase in RMSE due to using
\eqref{eq:opt} instead of \eqref{eq:optc}.
Due to the imperceptible difference between \eqref{eq:opt} and \eqref{eq:optc},
and the higher computational complexity of the latter,
the formulation \eqref{eq:opt} appears suitable for speech denoising using OGS.

\medskip
\noindent
\textbf{Further comparisons.}
This speech denoising example is intended as an illustrative example
of OGS rather than state-of-the-art speech denoising.
However, to provide further context, 
the output SNR of several algorithms, with and without empirical Wiener post-processing (EWP), are summarized in Table~\ref{table:sealgs}.
The additional algorithms are:
spectral subtraction (SS) \cite{Berouti_1979},
the log MMSE algorithm (LMA)  \cite{Cohen2002},
the subspace algorithm (SUB) \cite{Hu2003},
and persistent shrinkage (PS) \cite{Siedenburg_2013_JAES}.
For SS, LMA, and SUB, we used the MATLAB software provided in Ref.~\cite{Loizou_2007}.
For PS, we used the software provided by the authors.\footnote{\url{http://homepage.univie.ac.at/monika.doerfler/StrucAudio.html}} 
Without EWP, BT achieves the highest SNR of 15.35 dB.
However, the BT has slightly audible artifacts as noted above, as does PS.
The artifacts of SUB and OGS are less audible.
The artifacts of SS and LMA are clearly audible.

EWP improves the SNR of SUB and LMA by 1.14 dB and 1.83 dB; however, perceptual artifacts are still clearly audible.
EWP improves BT by 0.4 dB, and has a minor impact on perceptual quality.
EWP improves SUB, PS and OGS by 2.09 dB, 2.14 dB, and 1.86 dB, and slightly reduces the audible artifacts.
We note that the form of EWP used in PS is a \emph{persistent} form introduced in \cite{Siedenburg_2013_JAES}.
With EWP, four methods (BT, SUB, PS, and OGS) achieve almost the same SNR
(with varying degrees of audible musical noise).
However, SUB has a high computational complexity due to eigenvalue factorization.

\begin{table}
	\centering
	\caption{
	\label{table:sealgs}
	Output SNR of several speech enhancement algorithms, with and without empirical Wiener post-processing (EWP).
	} 
	\begin{tabular}{@{}lll@{}}
		\toprule
		 & \multicolumn{2}{c}{SNR (dB)} \\
		    \cmidrule(l){2-3}
		Method       &  w/o EWP & EWP  \\
		\midrule
		spectral subtract.~(SS) \cite{Berouti_1979}  &   13.50  & 14.64     \\
		log MMSE alg.~(LMA) \cite{Cohen2002}  &   13.10  & 14.93      \\
		subspace alg.~(SUB) \cite{Hu2003}     &   13.68  & 15.77  \\
		block-thresh.~(BT) \cite{Yu_2008_TSP}   &   15.35  & 15.75         \\
		persistent shrink.~(PS) \cite{Siedenburg_2013_JAES}  & 13.63 & 15.77 \\
		OGS (proposed)    &   13.77  & 15.63          \\
		\bottomrule
	\end{tabular}
\end{table}

\section{Conclusion}

This paper introduces a computationally efficient algorithm for denoising signals with 
group sparsity structure.\footnote{MATLAB software to implement the OGS algorithm and to reproduce the figures in the paper are online at \url{http://eeweb.poly.edu/iselesni/ogs/}.}
In this approach, `overlapping group shrinkage' or OGS, the groups are fully overlapping and the algorithm is translation-invariant.
The method is based on the minimization of a convex function.

A procedure is described for the selection of the regularization parameter $\lambda$.
The procedure is based on attenuating the noise to a specified level without
regard to the statistical properties of the signal of interest.
In this sense, the procedure for setting $\lambda$ is not Bayesian;
it does not depend on the signal to be estimated.
Even though the described procedure for setting $\lambda $ is conceptually simple,
it does not admit the use of explicit formulas for $\lambda$ because,
in part,
the OGS function does not itself have an explicit formula (it being
defined as the solution to a minimization problem).
The procedure to set $\lambda$ is based on analyzing the output
of OGS when it is applied to an i.i.d.\ standard normal signal,
and can be implemented in practice by off-line computation of 
tables such as Tables~\ref{tab:lamreal} and \ref{tab:lamcomplex}.
Adopting recent approaches for regularization parameter selection 
\cite{Deledall_2012_techrep} provides another potential 
approach to be investigated.

The paper illustrates the use of overlapping group shrinkage for speech denoising.
The OGS algorithm is applied to the STFT of the noisy signal.
Compared to the block thresholding algorithm \cite{Yu_2008_TSP},
the OGS algorithm gives similar SNR when Wiener post-processing is used.
However, the OGS denoised speech has fewer perceptual artifacts.

Current work includes the investigation of the group-sparsity promoting
penalty function proposed in \cite{Obozinski_2011_techrep, Jacob_2009_cnf},
namely, the derivation of a simple quadratic MM algorithm implementing
the proximal operator, the development of a procedure
for setting the regularization parameter $\lambda$ along the lines considered here, 
comparison of convergence behavior with other algorithms, 
and an evaluation for speech denoising.

\bibliographystyle{plain}


\clearpage

\onecolumn

\section*{Supplementary Material}

\noindent
MATLAB program for the implementation of overlapping group shrinkage (1D).

\prog{ogshrink.m} 

\newpage

\noindent
MATLAB program for the implementation of overlapping group shrinkage (2D).

\prog{ogshrink2.m}

\end{document}